# TuPy-E: detecting hate speech in Brazilian Portuguese social media with a novel dataset and comprehensive analysis of models

**Felipe Oliveira*, Victoria Reis, Nelson Ebecken**
Alberto Luiz Coimbra Institute of Post-graduation and Engineering Research
Federal University of Rio de Janeiro
felipe.oliveira@coc.ufrj.br

Social media has become integral to human interaction, providing a platform for communication and expression. However, the rise of hate speech on these platforms poses significant risks to individuals and communities. Detecting and addressing hate speech is particularly challenging in languages like Portuguese due to its rich vocabulary, complex grammar, and regional variations. To address this, we introduce TuPy-E, the largest annotated Portuguese corpus for hate speech detection. TuPy-E leverages an open-source approach, fostering collaboration within the research community. We conduct a detailed analysis using advanced techniques like BERT models, contributing to both academic understanding and practical applications.

## 1. Introduction

Social media has become a vital part of people's lives, offering popular means of communication and connection, allowing users to share information, experiences, and opinions in real time. While it fosters social connections, it also enables aggressive communication due to the anonymity it provides. As a result, hate speech and aggressive behaviors have become prevalent on social media platforms, posing risks to individuals and communities. Detecting and addressing hate speech is crucial for creating a safe online environment.

Particularities of the Portuguese language also add complexity to hate speech detection. With a rich vocabulary of over 600 thousand words and complex grammar, Portuguese may hide hate speech among numerous non-offensive words. Additionally, dialectal variations and regional expressions present challenges, as certain phrases may carry offensive meanings in specific contexts.

Inspired by the above justifications, we present the TuPy-E dataset, the largest public, and annotated corpus in Portuguese for automatic detection of hate speech. With an open-source approach, TuPy-E provides accessibility to the research community, encouraging collaboration and advancements in hate speech detection methodologies for Portuguese. The subsequent sections will delve into a detailed data analysis of TuPy-E, leveraging advanced techniques such as BERT models. This examination includes machine learning classifications for binary hate (hate or not) and hate categories, contributing to academic understanding and practical applications for model training and evaluation. Beyond its academic contributions, TuPy-E plays a pivotal role in developing more robust and culturally sensitive tools, contributing to a safer and more inclusive online environment [1]–[3].

In the subsequent section 2, we concisely overview the key related work—section 3 delves into the development of the TuPy-E dataset, outlining all the methodology steps; sections 4 and 5 comprehensively present TuPy-E corpus statistics, data analysis, and experimental results. Finally, Section 6 encapsulates concluding remarks and discussions.

## 2. Hate speech datasets and challenges

One of the significant challenges in automatically detecting hate speech is the need for publicly available annotated datasets. Added to this is that most of them are written in English. Here are some highlights from the literature:

A dataset of 16,914 tweets containing hate speech was created by [4] to study users who employ similar language patterns online. This dataset included 3,383 tweets related to misogyny, 1,972 related to racism, and 11,559 without hate speech content, supported by 613, 9, and 614 users, respectively. The dataset was manually curated by searching for tweets with slurs and terms linked to sexual, religious, gender, and ethnic minorities, even when they were not necessarily offensive. However, a limitation is that the tweet text can only be accessed through the public Twitter API [4].

The dataset described by authors [5] comprises 24,783 tweets that were manually categorized into hate speech with 1,430 instances, offensive but not hate speech with 19,190 instances, and neither hate speech nor offensive speech with 4,163 instances by two members of Figure-Eight. These tweets were collected via the Twitter API by filtering for tweets containing hate speech words sourced from Hatebase.org. This process yielded a sample of 33,548 instances from a pool of 85.4 million tweets collected from various user accounts. However, this dataset exhibits limited diversity in terms of hate speech content. Notably, gender-based hate speech tweets are skewed towards targeting women, and most of them focus on ethnicity-related content [5].

Researchers de Gibert et al. (2018) proposed a dataset constructed using content from Stormfront, the largest online community of white nationalists. This dataset focuses on discussions related to race, encompassing varying degrees of offensiveness. Annotation was performed at the sentence level to identify the minor units containing hate speech while minimizing noise. It includes 10,568 sentences categorized into hate speech (1,119 comments) and non-hate speech (8,537 comments). Additionally, two supplementary classes exist: "relation" for sentences expressing hate speech only when related to each other and "skip" for sentences not in English or lacking meaningful information for classification. The dataset also records metadata such as post-identifiers, sentence positions within posts, user identifiers, sub-forum identifiers, and the number of preceding posts reviewed by annotators. These samples were randomly selected from 22 subforums, providing diversity in topics and nationalities covered [6].

In a recent work by Mollas et al. (2020), the "ETHOS" dataset (multi-labEl haTe speecH detectiOn dataSet) was introduced. This dataset comprises textual data with two variations: binary and multi-label. It was compiled by collecting textual content from comments on YouTube and Reddit and later validated through the Figure-Eight crowdsourcing platform. The main objective of this research was to address a common challenge found in existing academic datasets: label imbalance. Specifically, the dataset consisted of 55.61% comments without hate speech and 44.39% comments containing hate speech content, which is a very sparse arrangement in this type of problem. Furthermore, it is worth noting that the text samples for evaluation came from unconventional platforms, such as Reddit and YouTube [7].

As discussed previously, the predominant focus of annotated datasets for hate speech detection has been the English language. However, annotated datasets are increasingly available in several other languages. Several noteworthy examples include:

The researchers compiled a corpus of annotated data from Facebook and Twitter for the French language. This corpus aids in the identification of instances of Islamophobia, misogyny, homophobia, religious intolerance, and ableism [8]. In the context of the German language, a corpus has been established to detect discrimination against immigrants and foreigners. This corpus encompasses 5,836 Facebook pages, where posts have undergone meticulous annotation and categorization. The posts encompass explicit and offensive messages targeting various subjects, including foreigners, governments, the press, and communities [9].

A recent multilingual research effort by [10] proposed a trilingual dataset encompassing English, French, and Arabic tweets. The primary aim was to identify variations in the expressions of 15 common phrases across these languages, specifically uncovering nuances within obscene phrases, particularly within sensitive discussions tied to specific geographical regions. To address linguistic intricacies within each language, a stringent rule set was imposed on human annotators recruited from the Amazon Mechanical Turk platform to ensure feedback reliability. An initial test set was provided for evaluation, and subsequent iterations refined the label set. Ultimately, this iterative process led to the creation of a dataset comprising 5,647 English tweets, 4,014 French tweets, and 3,353 Arabic tweets, each annotated across five distinct tasks. Although the single-task language models effectively handled binary classification for each tweet, the remaining four classification tasks, which included at least five label gradations, exhibited significant improvements when employing multitask models in both single- and multilingual settings.

Researchers Madhu et al. (2023) proposed the ICHCL dataset for the Hindi language. The authors consider that traditionally, research has treated hate speech recognition as a text classification problem, where algorithms determine whether a message is hateful based solely on that message's text. However, the work argues that context, or the messages and conversations tangential to the tweet, is crucial to understanding and addressing hate speech. In particular, the authors emphasize that short messages, such as tweets, can be easily misinterpreted if we do not consider previous messages in the conversation. The work extends previous efforts to classify hate speech by considering the current message and previous messages in a conversation [11].

In the context of the Arabic language, researchers Abdelhakim et al. (2023) proposed the Ar-PuFi dataset. Ar-PuFi is a significantly voluminous Arabic dataset consisting of 24,071 comments aimed at public figures in the Arab community, collected from TV interviews with Egyptian celebrities across various domains. Native speakers annotated the dataset to identify explicit and implicit offensive content, resulting in two- and six-class annotations. Several text representations were employed, with AraBERT as the base model. The dataset facilitated the development of lexicons of offensive terms associated with specific domains. The study also explored active learning and meta-learning to reduce labeling efforts, highlighting challenges in adapting Arabic dialects across different domains [12].

Furthermore, languages such as Greek, Croatian, and Indonesian have also seen the development of a dedicated corpus to facilitate discrimination detection [13]–[17].

In the context of the Portuguese language, four notable contributions stand out: Fortuna et al (2019) introduced a pioneering corpus started in 2017, comprising 5,668 annotated tweets covering European and Brazilian Portuguese. Subsequent updates have been made to accommodate the evolving lexicon in European Portuguese.

Pelle & Moreira (2017) released a corpus comprising 1,250 comments extracted from the Brazilian online newspaper G1, focusing specifically on Brazilian Portuguese. The authors performed annotations based on binary classes (offensive and non-offensive comments). Further, they categorized them into seven distinct hate speech groups, including racism, misogyny, homophobia, xenophobia, religious intolerance, and insults.

Leite et al. (2020) presented the ToLD-Br (Toxic Language Dataset for Brazilian Portuguese), an extensive dataset covering 21,000 tweets in Brazilian Portuguese manually annotated into seven categories: non-toxic, LGBTphobia, obscene, insult, racism, misogyny and xenophobia. The study employed several preprocessing techniques, including Bag of Words (BoW), Bidirectional Encoder Representations of Transformers (BERT), and AutoML techniques. Leite et al. (2020), in turn, carried out a multilingual analysis, exploring the applicability of toxic language detection models in languages other than Brazilian Portuguese, including English and Spanish.

Vargas et al. (2022) presented the second-largest dataset for detecting annotated hate speech in Brazilian Portuguese. Known as the HateBR corpus, this dataset was assembled from the comment sections of Brazilian politicians' Instagram accounts and meticulously annotated by experts in the field. The corpus presents three distinct layers of categorization: binary classification (offensive and non-offensive comments), offense level classification (highly, moderately, and slightly offensive), and nine distinct classes of hate speech, covering xenophobia, racism, homophobia, misogyny, religious intolerance, partisanship, endorsement of dictatorship, antisemitism, and fatphobia.

### 2.1. TuPy-E dataset creation

To overcome the notable shortcomings in existing Portuguese repositories of hate speech instances, we present the TuPy-E dataset. Recognizing the importance of prior research in this domain and the absence of annotated datasets for automated hate speech detection, we propose consolidating this comprehensive dataset by integrating the discoveries from Fortuna et al. (2019); Leite et al. (2020); Vargas et al. (2022), alongside a new, proprietary dataset.

Regarding the unpublished part of the TuPy-E dataset, we spent about seven months, from March 2023 to September 2023, building the corpus. We collaborated with a team of experts, including a linguist, a human rights lawyer, several behavior psychologists with master's degrees, and NLP and machine learning researchers.

A framework inspired by Vargas et al. (2022) and Fortuna (2017) was adhered to by establishing a stringent set of criteria for the selection of annotators, encompassing the following key attributes:

i) Diverse political orientations, encompassing individuals from the right-wing, liberal, and far-left spectrums.

ii) A high level of academic attainment comprising individuals with master's degrees, doctoral candidates, and holders of doctoral degrees.

iii) Expertise in fields of study closely aligned with the focus and objectives of our research.

The subsequent table provides a concise summary of the annotators' profiles and qualifications (Table 1).

Table 1 – Annotators' profiles and qualifications.

| Annotator | Gender | Education | Political | Color |
|---|---|---|---|---|
| #1 | Female | Ph.D. Candidate in civil engineering | Far-left | White |
| #2 | Male | Master's candidate in human rights | Far-left | Black |
| #3 | Female | Master's degree in behavioral psychology | Liberal | White |
| #4 | Male | Master's degree in behavioral psychology | Right-wing | Black |
| #5 | Female | Ph.D. Candidate in behavioral psychology | Liberal | Black |
| #6 | Male | Ph.D. Candidate in linguistics | Far-left | White |
| #7 | Female | Ph.D. Candidate in civil engineering | Liberal | White |
| #8 | Male | Ph.D. Candidate in civil engineering | Liberal | Black |
| #9 | Male | Master's degree in behavioral psychology | Far-left | White |

To consolidate data from the prominent works in the domain of automatic hate speech detection in Portuguese, we established a database by merging labeled document sets from Fortuna et al. (2019); Leite et al. (2020); Vargas et al. (2022). To ensure consistency and compatibility in our dataset, we applied the following guidelines for text integration:

i) Fortuna et al. (2019) constructed a database comprising 5,670 tweets, each labeled by three distinct annotators, to determine the presence or absence of hate speech. To maintain consistency, we employed a simple majority-voting process for document classification.

ii) The corpus compiled by Leite et al. (2020) consists of 21,000 tweets labeled by 129 volunteers, with each text assessed by three different evaluators. This study encompassed six types of toxic speech: homophobia, racism, xenophobia, offensive language, obscene language, and misogyny. Texts containing offensive and obscene language were excluded from the hate speech categorization. Following this criterion, we applied a straightforward majority-voting process for classification.

iii) Vargas et al. (2022) compiled a collection of 7,000 comments extracted from the Instagram platform, with three annotators labeled. These data had previously undergone a simple majority-voting process, eliminating the need for additional text classification procedures.

After completing the previous steps, the corpus was annotated using two different classification levels. The initial level involves a binary classification, distinguishing between aggressive and non-aggressive language. The second classification level involved assigning a hate speech category to each tweet previously marked as aggressive in the previous step. The categories used included ageism, aporophobia, body shame, capacitism, LGBTphobia, political, racism, religious intolerance, misogyny, and xenophobia. It is important to note that a single tweet could fall under one or more of these categories.

The following section will provide a comprehensive description of the methodology employed.

## 3. Methodology

Initially introduced in 2006, Twitter emerged as a micro-blogging platform, enabling individuals to disseminate concise, 140-character messages, commonly called "tweets."

In the user count ranking, Twitter secures 14th, trailing behind platforms like Facebook, Instagram, and TikTok. Interestingly, a mere 10% of Twitter's user base contributes to 92% of the tweets. Most Twitter users exhibit low levels of activity in terms of content sharing. Most users only engage in tweeting once per month on average. This implies that most individuals come to Twitter primarily to consume content rather than create it. Data demonstrates that a distinct, highly active subset of users, constituting 10% of the total, generates a substantial portion of the platform's content.

Twitter has 396.5 million global users, with a daily volume of at least 500 million tweets. Brazil contributes significantly, with 19.05 million active Twitter accounts. To advance with the study, a sequence of procedures was implemented. The subsequent diagram presents the process schematics, and further elaboration will be provided in the subsequent sections of this chapter.

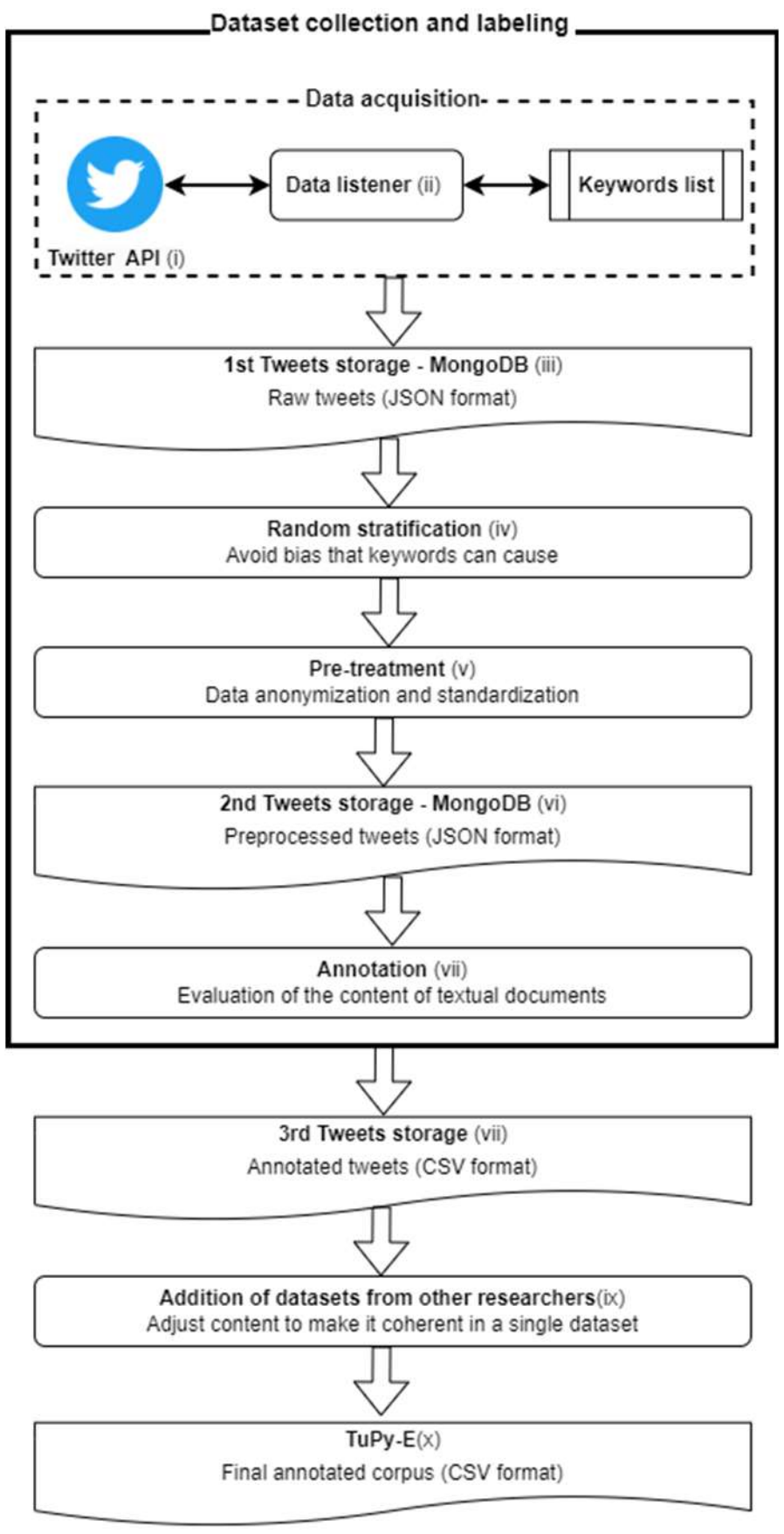

Figure 1 - The proposed approach for the TuPy-E dataset construction.

### 3.1. Tweets collection

Following the schema in Figure 1, it is essential to begin by explaining how the Twitter API was used during the acquisition phase of this work. Note that recent changes have significantly altered the politics described in this work.

Twitter's Developer Platform provides real-time and historical Twitter data access tailored for various purposes, including business, non-profit initiatives, educational endeavors, and research pursuits. To utilize the Twitter Developer Platform, the user must create an account, either an existing personal Twitter account or a dedicated one specifically for development purposes.

Access to Twitter data via the Developer Platform is categorized into different levels, one of which is known as Academic Research access. This particular level offers academic researchers' free access to real-time data and extensive archival search capabilities. To gain Academic Research access, users must undergo an application process, during which Twitter evaluates their academic qualifications, research objectives, methodologies, and strategies for disseminating knowledge. This level is accessible to graduate students, doctoral candidates, postdoctoral fellows, faculty members, and individuals engaged in research activities associated with academic institutions or universities.

Applicants must furnish additional information beyond their Twitter user profile, including their institution-registered name, a web page for identity verification, particulars about their university, department, school, or laboratory, field of study, and current professional role. In terms of their research project, applicants must provide comprehensive details about their proposed research, encompassing research questions, funding sources, analytical methods, and plans for sharing their research findings.

In conjunction with the Academic Research access described earlier, the Twitter API offers searchable access to all public tweets generated within the last seven days in real-time, along with Twitter's comprehensive public archive dating back to March 2006.

For the data acquisition (steps i and ii of Figure 1), we opted to utilize a design pattern known as the "listener" pattern, enabling the definition of an update mechanism that notifies multiple objects about any events occurring with the observed object (in our case, new tweets). This step was implemented using the Python 3.10.0 programming language and aligns with the typical streaming design approach.

We employ the widely popular Python library, Tweepy, known for its easy-to-use and efficient features in interacting with the Twitter API to achieve this goal. Tweepy streamlines the authentication process, simplifies request handling, and streamlines response management. This library is invaluable to many professionals, including developers, data scientists, and researchers, who seek to access and analyze Twitter data for many applications seamlessly.

We implemented a 15-minute update pattern for our tweet collection process from May to July 2023. This process involved selecting batches of tweets based on three distinct criteria: i) identifying texts containing specific keywords and terms, as L. Silva & Sampaio (2017) defined (Table A.1 with words and expressions in the attachment section), ii) flagging texts with offensive content, as determined by the authors of this study, and iii) random selection of texts without any prior filtering. It is important to emphasize that a fundamental hypothesis in our research, focused on the spread of hate speech in Brazil, revolves around political connotations of hate. This led us to develop filters focused on this aspect, including insights from the work of [21], the Brazilian study most related to this issue.

Tweets were sourced from subcollections known as dictionaries, which are structured information repositories providing context and details for each specific tweet. These details include the tweet ID, creation date, text content, user information, retweet count, and more.

The consolidation process, depicted in step iii of Figure 1, was performed using MongoDB, a renowned database software for storing JSON-like objects, including tweets. MongoDB is a popular NoSQL database management system designed to handle substantial volumes of data, particularly unstructured or semi-structured data. It operates as an open-source, document-oriented database using a flexible, schema-less data model that makes it suitable for various applications, including ours.

### 3.2. Tweets pre-treatment

This stage can be summarized as preparing the data for the annotation phase. The initial preprocessing involves a stratified sampling of reality to represent the imbalanced nature of the phenomenon the dataset seeks to study. By acquiring the dataset using the three criteria mentioned in the previous section, we have reduced the raw data size while maintaining the proportion between the collected tweets with the specified terms and keywords and the unfiltered tweets. This way, we avoid the risk of structuring data with only one or no terms.

Scikit-learn, commonly abbreviated as Sklearn, was used to address this matter. Sklearn is a Python library designed for machine learning and data science. More specifically, the Stratified k-fold argument was applied, ensuring that each subset contains a proportionate representation of samples from every target class, maintaining the same percentage as the complete dataset (step iv in Figure 1).

Another action taken regarding the massive volume of data acquired in the acquisition process was the removal of records related to retweets. When a user decides to retweet a tweet, a new data entry is generated in the database to document this action. Typically, this entry contains information about the user who performed the retweet (referred to as the "retweeter"), the ID of the original tweet being retweeted, the timestamp of the retweet, and any additional metadata, such as the user's comments or added text. In this context, all duplicated text resulting from retweets was eliminated from the database.

Another crucial action taken to carry out this task was anonymizing the users of the collected tweets. As previously mentioned, one of the primary identifiers within the collected data is the user, whose representation typically includes the "@" symbol followed by their chosen username on the social media platform. Utilizing this structure, all text blocks adhering to this format were meticulously removed from the processed database. In addition, all external links within the text were deleted, and all text was converted to lowercase (step v in Figure 1).

Finally, in step vi of Figure 1, a second consolidation in MongoDB has been executed, now with a more suitable data volume for the annotation process and including more structured and anonymous data.

### 3.3. Tweets annotation

As Vargas et al. (2022) pointed out, the definitions of offensive language and hate speech must be delineated. They are often used within the exact overarching definition of toxic language, as seen in the study by Gaydhani et al. (2018). This overlapping usage can lead to challenges in classification tasks.

The definition of hate speech needs to be more uniform and straightforward, as numerous entities have put forth their interpretations and guidelines. According to Nobata et al. (2016), hate speech encompasses any language that targets, degrades, and incites violence against a group based on their race, ethnicity, origin, religion, disability, gender, age, or sexual orientation/gender identity. Establishing a clear, unified definition is essential for its automation through machine learning.

In her research, Fortuna (2017) lists various definitions, not only from government entities and the previously mentioned Nobata et al. (2016) but also from the terms of use of social media platforms such as YouTube, Facebook, and Twitter. These companies introduce nuances in their conceptualizations, such as the role of sarcasm and humor as contributing factors. For instance, YouTube mentions a "fine line" between what is or is not considered hate speech. For example, it is generally acceptable to criticize a nation-state. However, it is not appropriate to post hateful or malicious comments about a group of people solely based on their ethnicity.

Building upon this, the definition attributed in this work is that hate speech is the use of language that attacks or degrades, incites violence, or promotes hatred against groups based on specific characteristics such as physical appearance, religion, national or ethnic origin, sexual orientation, gender identity, among others, and this can manifest in various linguistic styles, even in subtle forms or when humor is employed. Meanwhile, offensive language cases were characterized by the utilization of abusive expressions, threats, and derogatory terms.

A representative example of this difference is the commonly used and hostile expression in Brazilian Portuguese, "vai se f****," which is a profane way of telling someone to "go f*** themselves" or "go to hell." It is a solid and offensive phrase, but it does not qualify as hate speech because it does not target any specific group.

As discussed in section 2.4 of this work, the evaluation of tweets followed a two-tiered process. In the first tier, the evaluator made a binary classification, assigning tweets to one of two categories: aggressive or non-aggressive. Following this assessment, the presumption is that all data identified as hate speech are included within the group labeled as "aggressive" in the first classification layer.

The second layer involves assigning the hate speech categories to the previously classified text as offensive. The categories chosen for this study include ageism, aporophobia, body shame, capacitism, LGBTphobia, political, racism, religious intolerance, misogyny, and xenophobia.

It is important to note that a single text can contain more than one type of hate speech, as in the example: "@user @user said it all, my friend, the poor on the right should all go to hell," in which the hostile language is classified as aporophobia (disdain for poor people) and political hatred (disdain for people following a particular political ideology). This step in the process aligns with step vii in Figure 1.

### 3.4. Dataset characteristics

The TuPy-E[1] dataset results from the amalgamation of the combined datasets from Fortuna et al. (2019); Leite et al. (2020) and Vargas et al. (2022), along with a new, unpublished dataset created during this study. Categorically, "non-aggressive" makes up the majority at approximately 72.08%, with "Aggressive" accounting for roughly 27.91%. Additionally, "Not Hate" represents approximately 79.15% of the dataset, while "Hate" comprises about 20.84%. The dataset consists of 43,668 instances, with these percentages describing the label distribution. The totals are described in Table 2.

Table 2 – Count of tweets for categories non-aggressive and aggressive

| Label | | Count |
|---|---|---|
| Non-aggressive | | 31121 |
| Aggressive | Not hate | 3180 |
| | Hate | 9367 |
| Total | | 43668 |

Table 3 presents a taxonomy of hate labels and the corresponding percentages based on the provided data. The categories include "ageism" (0.61%), "aporophobia" (0.71%), "body shame" (3.05%), "capacitism" (1.06%), "LGBTphobia" (8.61%), "political" (12.27%), "racism" (3.10%), "religious intolerance" (1.15%), "misogyny" (17.92%), "xenophobia" (3.81%), and a residual category labeled as "other" (47.81%).

The total dataset size is 9367, and it is essential to note that a single tweet can fall under multiple hate categories, as indicated by the percentages in the table.

Table 3 - Count of tweets for each category of hate

| Label | Count |
|---|---|
| Ageism | 57 |
| Aporophobia | 66 |
| Body shame | 285 |
| Capacitism | 99 |
| LGBTphobia | 805 |
| Political | 1149 |
| Racism | 290 |
| Religious intolerance | 108 |
| Misogyny | 1675 |
| Xenophobia | 357 |
| Other | 4476 |
| total | 9367 |

Table 4 provides valuable insights into the distribution patterns of "aggressive" and "hate" labels within the dataset, as observed by various researchers. Examining the results, Fortuna et al. reveal a lower incidence of aggression (11%) and a comparatively higher prevalence of hate (18%). On the other hand, Leite et al. identifies a substantial skew towards aggression, with a prominent 43%, while hate stands at 34%.

[1]https://huggingface.co/datasets/Silly-Machine/TuPyE-Dataset

The findings by Vargas et al. showcase a relatively balanced distribution, with 31% for aggression and 33% for hate.

Notably, the unpublished TuPy-E shows an equal prevalence of 15% for aggression and hate, suggesting a unique symmetry in the dataset. This table underscores the nuanced variations in the distribution of aggressive and hate-related content perceived by the different used research perspectives.

Table 4 – Proportions for labels aggressive and hate for each researcher.

| Researcher | aggressive % | hate % |
|---|---|---|
| Fortuna et al | 11 | 18 |
| Leite et al | 43 | 34 |
| Vargas et al | 31 | 33 |
| Unpublished | 15 | 15 |

Table 5 articulates the distribution of label categories as a percentage of the overall dataset count for each researcher. It discerns how various researchers categorize the dataset. Fortuna et al. ascribes the highest percentage to "misogyny" (47.6%) and "LGBTphobia" (19.1%). Leite et al. proffers a predominantly elevated percentage to the category "other" (80%) while allocating relatively minor percentages to most other categories. Vargas et al. distribute the percentages more uniformly across various categories, eschewing a prominent bias. The unpublished TuPy-E part assigns a considerable proportion (77.3%) to the category "political" while maintaining the lowest incidence in most other categories.

Table 5 – Proportions of each category of hate for researcher

| Category | Fortuna et al. | Leite et al. | Vargas et al. | Unpublished |
|---|---|---|---|---|
| Ageism | 0.2 | 0.0 | 4.1 | 0.0 |
| Aporophobia | 0.2 | 0.0 | 4.8 | 0.0 |
| Body shame | 8.6 | 0.0 | 9.3 | 0.0 |
| Capacitism | 0.3 | 0.0 | 7.2 | 0.0 |
| LGBTphobia | 19.1 | 6.3 | 7.4 | 0.0 |
| Political | 4.9 | 0.0 | 40.9 | 77.3 |
| Racism | 5.6 | 2.5 | 2.9 | 1.2 |
| Religious intolerance | 1.6 | 0.0 | 2.2 | 7.2 |
| Misogyny | 47.6 | 8.5 | 15.9 | 14.3 |
| Xenophobia | 7.1 | 2.8 | 5.4 | 0.0 |
| Other | 4.8 | 80 | 0.0 | 0.0 |

The unique dataset shows that the different researchers report varying percentages for the labels "aggressive" and "hate." This discrepancy can be attributed to researcher subjectivity in labeling. What one researcher may categorize as "aggressive" content, another might categorize differently. The same applies to the "hate" label. These variances reflect differences in the researchers' subjective judgments, interpretations, and the specific criteria they use for classifying content.

Various factors can influence this subjectivity, including individual backgrounds, perspectives, and the absence of universally accepted standards for defining and categorizing aggression and hate. It underscores the inherent challenges in achieving consistency and objectivity in content analysis, especially in sensitive topics where interpretations may be open to debate.

## 4. Data analysis

Table 6 overviews the top 10 most frequently occurring words within the broader hate dataset. The prevalence of terms like "mulher" (woman) underscores the notable emphasis on gender within the realm of hate speech. Disturbingly, offensive expressions like "gorda" (fat) and "feia" (ugly) not only signify instances of body shaming but also contribute to an evident gender bias, given their association with language commonly directed at females. This underscores a troubling pattern wherein hate speech consistently revolves around appearance, indicating a proclivity to target individuals, predominantly women, based on their physical attributes.

Table 6 – Count of the most frequent words of TuPy-E dataset

| Word | Frequency |
|---|---|
| Mulher | 380 |
| Cara | 271 |
| Cu | 260 |
| Feia | 219 |
| Caralho | 189 |
| Burra | 179 |
| Gorda | 178 |
| Porra | 174 |
| Homem | 167 |
| Brasil | 166 |

The “other” words, including explicit terms like "c******," "c*," and "p****," contribute to an overall offensive tone, emphasizing the repetitive use of strong language. Additionally, terms like "burra" (dumb, in feminine) extend derogatory language beyond appearance, targeting intelligence. The presence of "homem" (man) indicates a recurring gender theme, while "brasil" (Brazil) introduces a geopolitical aspect, hinting at offensive language directed at individuals associated with a specific political ideology. Together, these elements paint a comprehensive picture of the diverse, however biased, hate speech dataset.

Table A.2, in the attachment section, presents a breakdown of hate speech in the categories mentioned earlier, each associated with the top 5 specific words and their frequencies. Noteworthy terms and their frequencies are outlined, shedding light on the prevalence of specific themes within the dataset.

The occurrence of certain words in multiple categories within Table A.2 highlights the complex and interconnected nature of hate speech. Words that appear in more than one category suggest a differentiated layer of discriminatory language that intercedes with singular themes, reflecting the complexity of prejudices present in the data set and how they overlap.

For example, the word "woman" appears in both the misogyny and LGBTphobia categories, indicating a gender-specific term that is also used in the context of discrimination against the LGBTQ+ community. Likewise, the term "brasil" appears in both the political and xenophobia categories, suggesting that hate speech involves both political affiliations and prejudice against individuals associated with a specific region.

In the political context, the term "Lula" is associated with a specific political figure or ideology, indicating that hate speech extends beyond mere political disagreement to encompass personal attacks and prejudices. In the aporophobia category, "Lula" takes on an additional dimension, implying a discriminatory attitude towards socioeconomically disadvantaged individuals.

This analysis highlights the multifaceted nature of hate speech, showing that specific words function as conduits for prejudice against multiple social groups.

### 4.1. N-grams

N-grams are a fundamental and versatile concept in natural language processing and text analysis. N-grams are compact sequences of contiguous items within a given piece of text. These items can be as short as individual words or extend to pairs and trigrams of words, known as bigrams and trigrams. Exploring N-grams provides a nuanced approach to understanding the intricate structures and patterns that underlie language.

For this phase of analysis, the texts underwent tokenization, converting words to lowercase letters so as not to differentiate between uppercase and lowercase letters, and the subsequent removal of punctuation. Stopword elimination, facilitated by the NLTK library's predefined list and user-specified exclusions, refined the dataset.

Finally, the processed words are consolidated into a list to generate N-grams, creating a differentiated representation of linguistic structures from unigrams to trigrams. The ten most frequent N-grams are then plotted quantitatively, providing a technical perspective on the prevailing linguistic patterns in the text.

The corpus presents 40,363 unique words and covers various Brazilian and Portuguese idiomatic expressions. Figure 2 illustrates the ten most frequent bigrams in texts where hate speech was not identified.

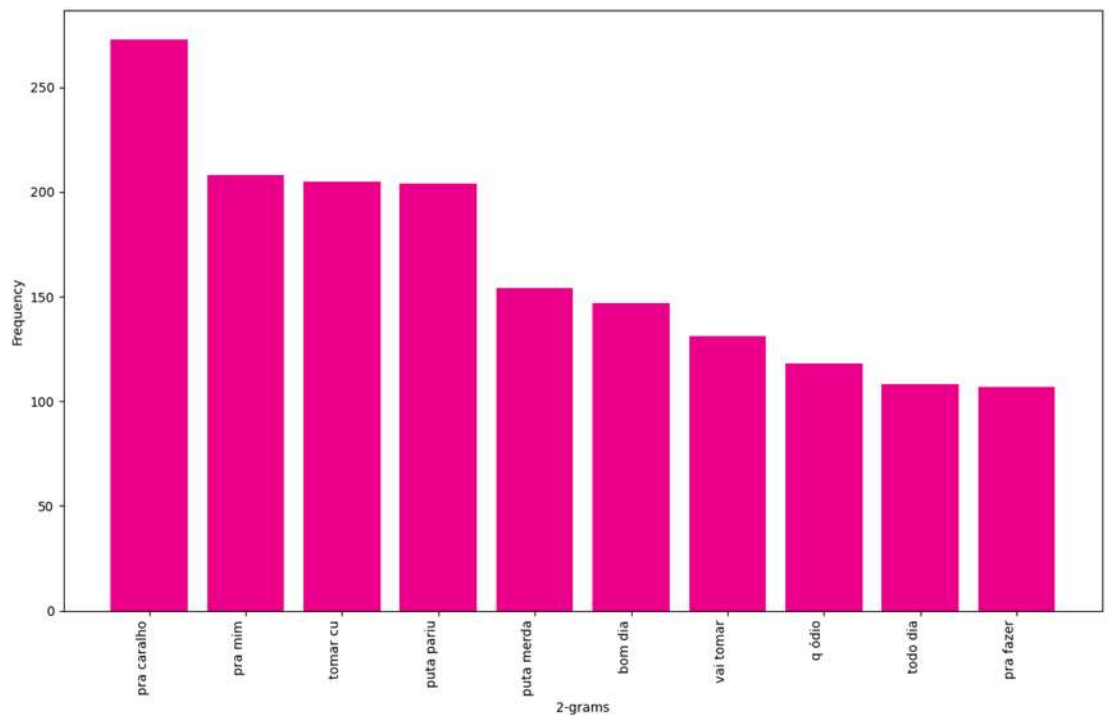

Figure 2 – Ten most frequent bigrams for not hate category.

In Figure 2, it is possible to observe the recurrence of offensive or obscene terms even in texts where hate speech was not identified, concluding that there is a trivialization of profanity and obscene words even when there is no hate speech. Figure 3 illustrates the ten most frequent bigrams in texts where hate speech was identified.

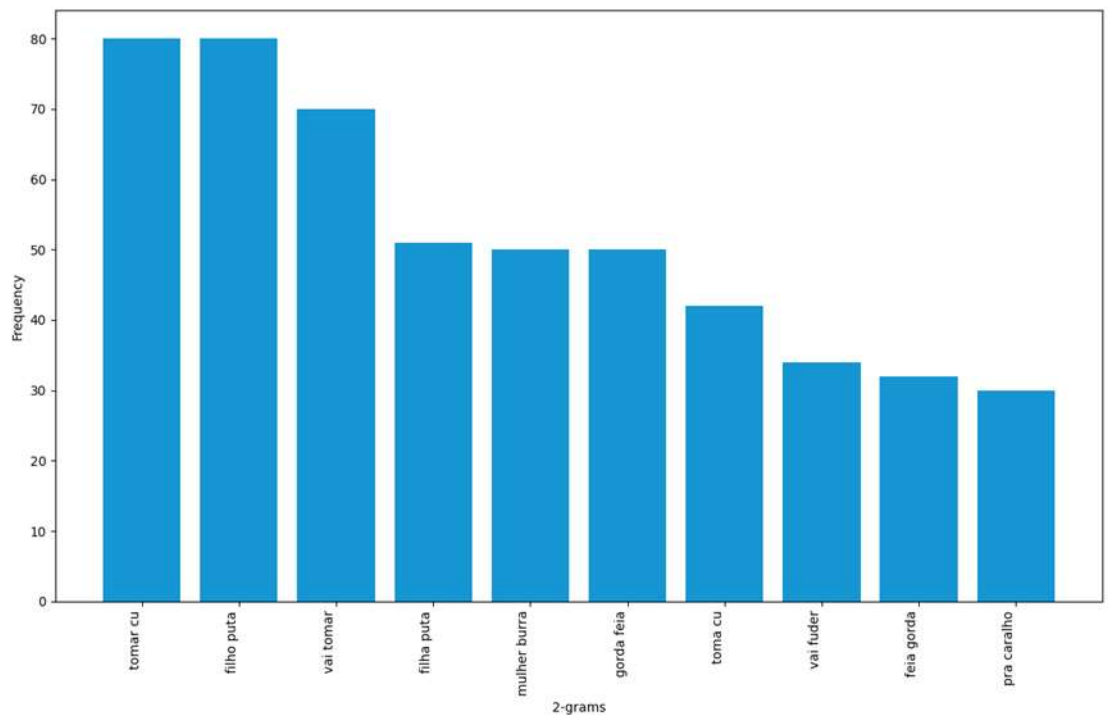

Figure 3 – Ten most frequent bigrams for the hate category.

The analysis of Bigrams reveals an interesting juxtaposition of both benign and obscene words in both hate and non-hate contexts. In the non-hate category, expressions such as “para me” (for me), “bom dia” (good morning), and “todo dia” (every day) are relatively neutral and commonly used in everyday conversations. On the contrary, the hate category introduces a strong contrast with bigrams such as "toma c*," "filho p***," "mulher burra," and "gorda feia," which involve explicit and offensive language directed at individuals based on ties, family members and physical appearance.

The presence of certain bigrams, such as “pra c******,” in both hateful and non-hateful contexts highlights their versatile use, but it is crucial to note the change in tone and intent when used in hate speech. The bigram “vai t****” exemplifies a similar duality, existing in both categories but taking on a more aggressive and explicit character in the context of hate.

This duality in the use of certain bigrams emphasizes the nuanced nature of language and the importance of context in deciphering meaning [22]. It also highlights the complexity of hate speech, where seemingly innocuous phrases in each context can turn into offensive expressions when used with malicious intent.

### 4.2. Graphs

A graph serves as a structured framework, representing information in a graph format where the entities are nodes and the relationships between them are edges. Each node and edge can have associated attributes, facilitating the organization and efficient information retrieval and promoting a semantic understanding of relationships. Graphs capture connections and meanings, making them valuable for AI, data integration, semantic search, and recommendation systems, and creating interconnected information networks.

To create the graphs for our dataset, we created a weight matrix of zeros, scaled by unique words in the dataset. Word co-occurrences were examined by splitting the text data, identifying pairs within a specified window size, and updating the weight matrix. The importance of words was determined by the sum of the weights, indicating the frequency of co-occurrence. The N most influential words are selected, and a NetworkX module represents a graph of words. The node's size reflects the word's relevance, and the edges indicate the strength of the co-occurrence.

Figure 4 shows the graph of texts marked as hate speech. Noticeably, all observations regarding the hate profile made in the bigrams are reaffirmed in the plotted graph. The most significant nodes continue to be associated with gender-biased insults and offensive language with connotations of political ideology. Thicker borders, representing the connections between words, depict pairs typically linked to hate, such as "filho p***," "mulher burra," and "tomar c*."

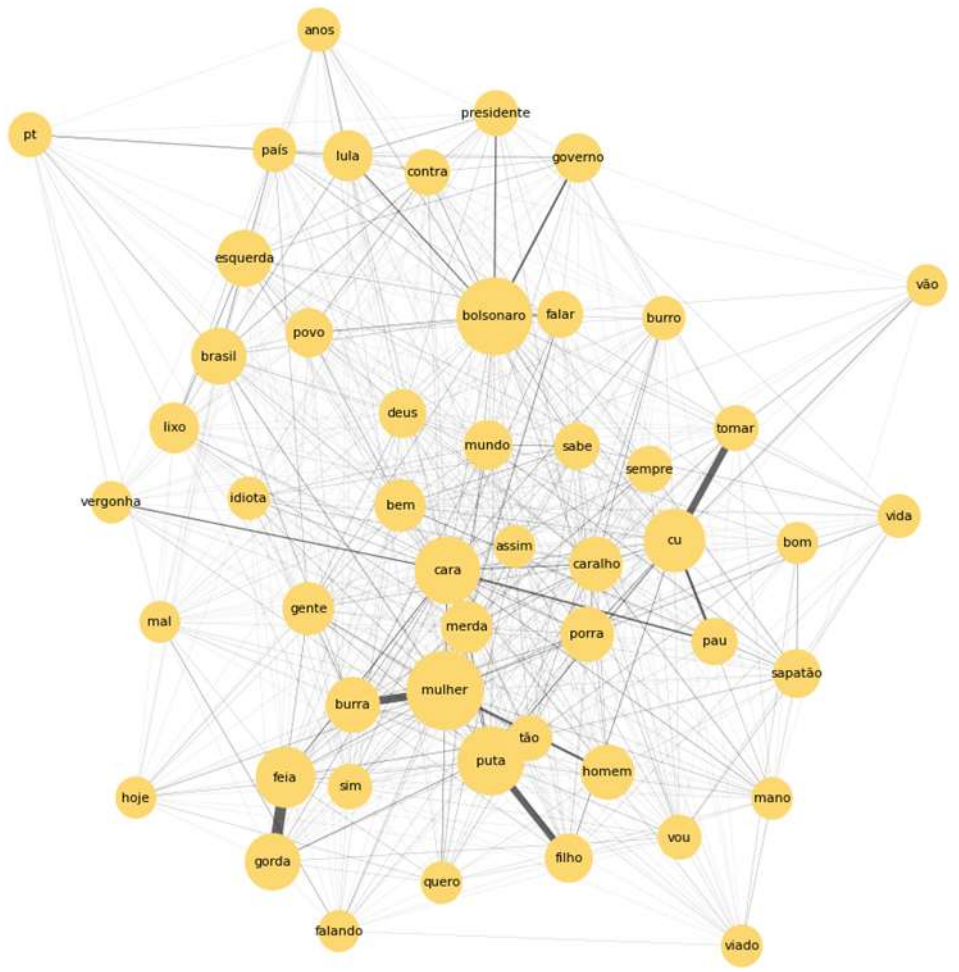

Figure 4 – Graph of words in hate

Figure 5 and Figure 6 delve deeper into the distinctive realms of misogyny and political bias, respectively. The nodes in these graphs shed light on the most prominent and interconnected terms within each category. The visual representation amplifies the weight these themes carry and showcases the specific words and phrases that contribute significantly to the discourse of hate within these spheres.

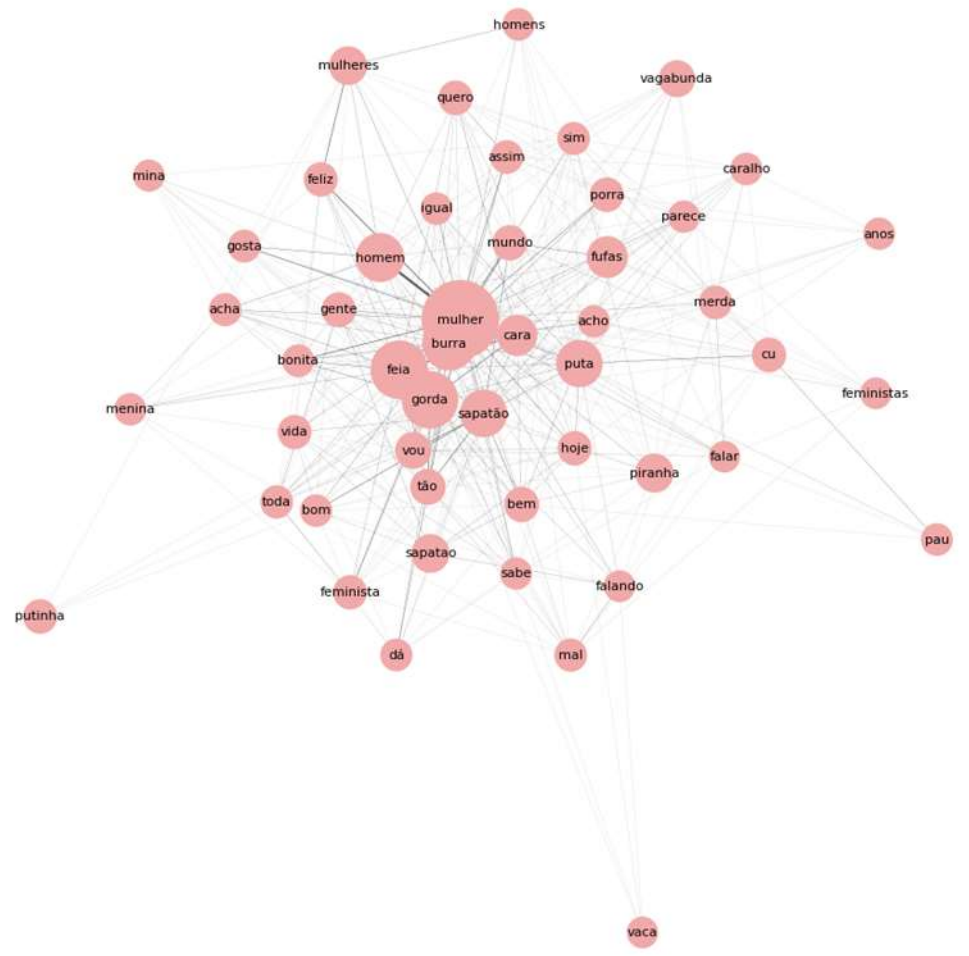

Figure 5 – Graph of words in hate in the category of misogyny

Upon a closer examination of Figure 5, it becomes evident that the graph does not exhibit fewer distinct clusters. Instead, a notable concentration is observed around a central node, particularly the term "woman." This singular focus on the word "woman" indicates a central theme within misogyny on Brazilian Twitter. The interconnectedness of terms surrounding "woman" suggests that derogatory language and gender-based insults are prominently linked, forming a cohesive subgraph. This focused structure underscores a prevailing narrative within misogyny, with the term "woman" acting as a central node that attracts and connects various expressions of hate speech.

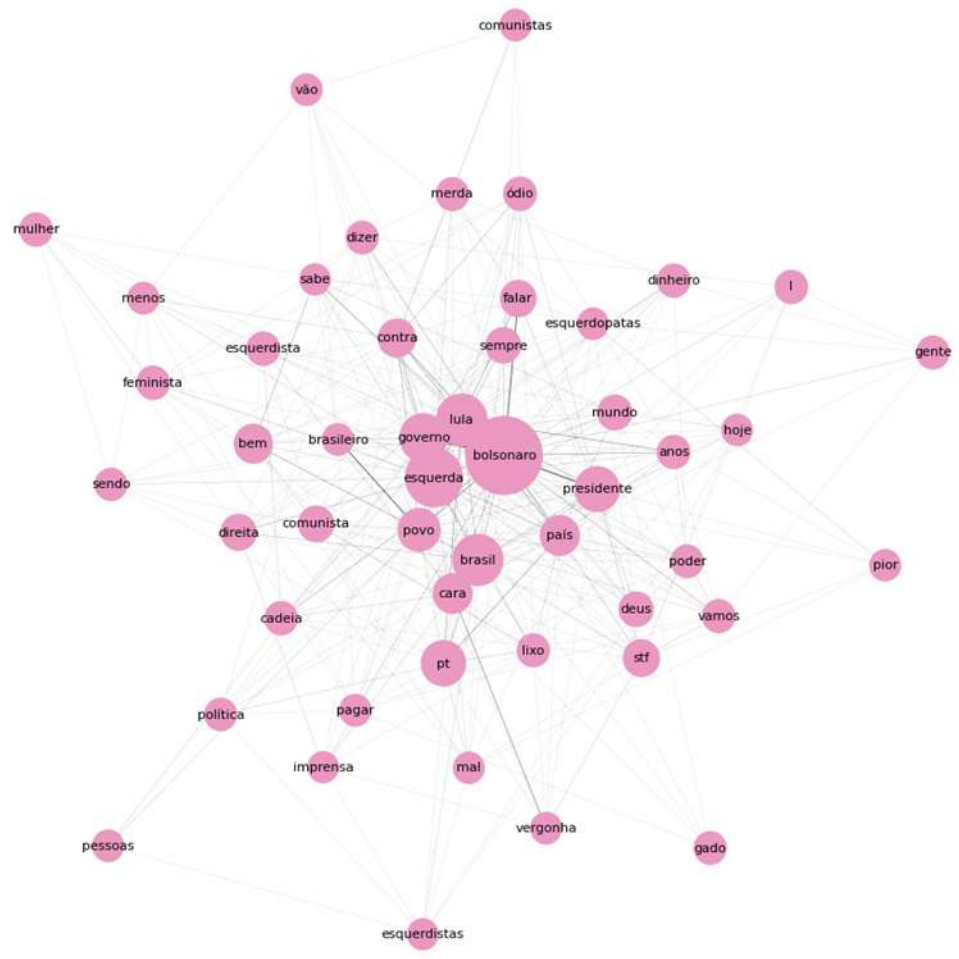

Figure 6 – Graph of words in hate in the category of political

In Figure 6, the political bias graph unveils a noteworthy observation about the shared use of words across the political spectrum. Terms associated with political figures, ideologies, and derogatory remarks are interconnected, forming cohesive clusters. Notably, some words are employed for candidates across the right and left political spectrum. This suggests that certain expressions and insults are not exclusive to a particular political orientation but are shared across ideological boundaries. The interconnected nodes related to political terms underscore the entwined nature of hate speech with political discourse, revealing a common lexicon that spans the political spectrum.

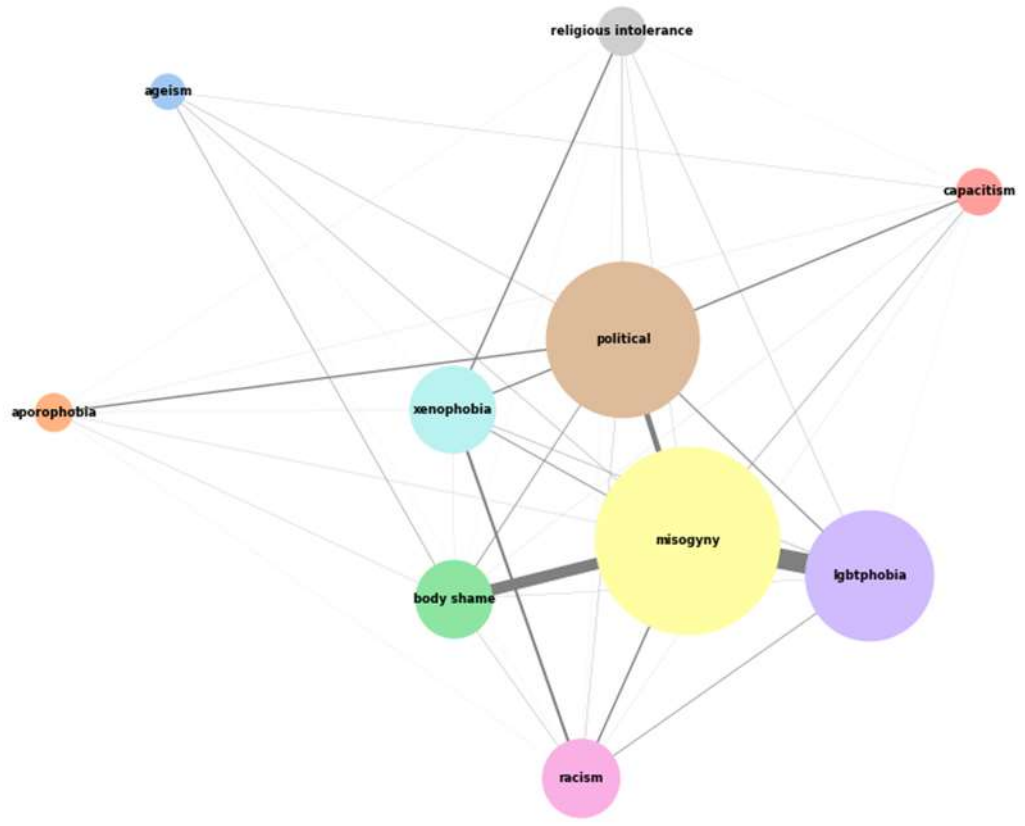

Figure 7 – Graph of hate categories.

Building upon the hypotheses generated from the previous analyses, which outlined a profile of hostile language in Brazilian Portuguese, we analyzed co-occurrence among the hate categories themselves (Figure 7). In other words, we observed which types of hate speech are more closely associated with each other. As expected, there is a very strong co-occurrence between misogyny, body shame, and LGBTphobia, with considerable connections to other categories as well—those that feature repeated words in their appearance rankings.

Regarding politics, we observed connections between aporophobia and capacitism, as mentioned in Section 4 of this study. Additionally, there is a proximity between racism, xenophobia, and religious intolerance, which, due to their lower occurrence, may not have been evident in previous analyses but emerged in the graphical analysis.

## 5. Models

The dataset's prior goal is to be instrumental in advancing artificial intelligence, specifically in the identification of offensive language. Through a detailed assessment of various benchmark classifiers, we aimed to identify the most effective predictor within the TuPy-E dataset. The primary focus is conducting an exhaustive analysis of offensive language itself and determining whether discernible patterns in word usage are distinct enough for accurate predictions.

Two versions of the BERTimbau Cased model were fined tuned using TuPy-E corpus for two distinct classification tasks: binary classification for hate, and hierarchical classification (categorical hate). The following section will provide a more detailed description of the models used [23].

### 5.1. BERTimbau

BERTimbau is a pre-trained BERT model designed for Brazilian Portuguese and stands out as a breakthrough in natural language processing. Built on the Bidirectional Encoder Representations from Transformers (BERT) architecture, this model underwent extensive pre-training on a vast corpus of Brazilian Portuguese text data. The result is a nuanced understanding of contextual information and semantic relationships within language.

Available in two configurations, Base and Large, the BERTibau offers adaptability for various applications. The Large variant, characterized by greater complexity, can perform better but requires increased computational resources, by using 334 million parameters.

The practical applications of BERTimbau Base cover information extraction, document understanding, and sentiment analysis in Brazilian Portuguese texts. This variant uses 109 million parameters.

Within the broader landscape of NLP research, the open-source availability of BERTimbau Base serves as a catalyst, allowing researchers and developers to leverage its pre-trained capabilities or tune it for specific tasks and domains.

### 5.2. Binary classification for hate detection

In the context of hate speech detection, the performance of two pre-trained language models, BERTimbau Base[2] and BERTimbau Large[3], was evaluated. The objective was to discern their effectiveness in identifying instances of hate speech within Brazilian Portuguese content. A fine-tuning method was employed to adapt the pre-trained model for TuPy-E dataset.

Due to their difference in processing, we applied ten and twenty epochs for the variants large and base, respectively. The dataset was sliced in 80% train and 20% test. Results are shown in Table 7.

Table 7 – Metrics for BERTimbau Base and BERTimbau Large models for binary prediction.

| Model | Accuracy | Precision | Recall | F1 Score |
|---|---|---|---|---|
| BERTimbau Base | 0.901 | 0.897 | 0.901 | 0.899 |
| BERTimbau Large | 0.907 | 0.901 | 0.907 | 0.903 |

Both models exhibited commendable accuracy, achieving an impressive 90%, underscoring their proficiency in correctly classifying hate speech.

Notably, both models' precision and recall scores struck a balance, reflecting a judicious trade-off between minimizing false positives and false negatives. This equilibrium is crucial in detecting hate speech, preventing models from downplaying, or exaggerating the prevalence of offensive content. The consistent F1 score, accounting for precision and recall, further underscores the models' robustness as a comprehensive evaluation metric.

Surprisingly, both BERTimbau Base and BERTimbau Large exhibited remarkably similar performance across all metrics. This suggests that the base model performs comparably to its larger counterpart within the specific domain of hate speech detection. This finding is significant, especially considering the resource constraints for practical applications. The base model appears pragmatic, offering efficiency without compromising performance.

[2] https://huggingface.co/Silly-Machine/TuPy-Bert-Base-Binary-Classifier
[3] https://huggingface.co/Silly-Machine/TuPy-Bert-Large-Binary-Classifier
[4] https://huggingface.co/Silly-Machine/TuPy-Bert-Base-Multilabel
[5] https://huggingface.co/Silly-Machine/TuPy-Bert-Large-Multilabel

The robustness of these models is further highlighted by their high precision and balance between precision and recall. This adaptability is crucial for real-world applications, where models must navigate the diverse and evolving expressions of hate speech in language.

While overall performance is notable, drilling into model predictions by category can provide valuable insights. This detailed analysis is essential for refining model capabilities and enhancing their effectiveness in addressing the intricacies of hate speech detection in Brazilian Portuguese content.

## 5.3. Hierarchical classification

The hate speech detection results from two models, BERTimbau Base[4], and BERTimbau Large[5], reveal valuable insights into their performance across the various categories of hate.

Both models exhibit reasonable precision, recall, and F1 scores regarding overall effectiveness. However, a closer examination of specific hate speech categories uncovers nuances in their capabilities (see Table 8).

Table 8 – Metrics for BERTimbau Base and BERTimbau Large models for Hierarchical classification.

| Model | Category | Precision | Recall | F1 score | Support |
|---|---|---|---|---|---|
| BERTimbau Base | Ageism | 1.00 | 0.00 | 0.00 | 15 |
| | Aporophobia | 1.00 | 0.00 | 0.00 | 16 |
| | Body shame | 0.58 | 0.54 | 0.56 | 54 |
| | Capacitism | 1.00 | 0.00 | 0.00 | 20 |
| | Lgbtphobia | 0.85 | 0.67 | 0.75 | 171 |
| | Political | 0.59 | 0.56 | 0.58 | 220 |
| | Racism | 0.29 | 0.27 | 0.28 | 62 |
| | Religious intolerance | 0.25 | 0.11 | 0.15 | 19 |
| | Misogyny | 0.65 | 0.60 | 0.62 | 324 |
| | Xenophobia | 0.41 | 0.31 | 0.35 | 78 |
| | Other | 0.56 | 0.49 | 0.52 | 909 |
| | Not hate | 0.92 | 0.93 | 0.92 | 7177 |
| | **Micro avg** | **0.86** | **0.84** | **0.85** | **9065** |
| | **Macro avg** | **0.67** | **0.37** | **0.39** | |
| | **Weighted avg** | **0.85** | **0.84** | **0.84** | |
| | **Samples avg** | **0.86** | **0.85** | **0.85** | |
| BERTimbau Large | Ageism | 0.40 | 0.13 | 0.20 | 15 |
| | Aporophobia | 0.75 | 0.19 | 0.30 | 16 |
| | Body shame | 0.78 | 0.65 | 0.71 | 54 |
| | Capacitism | 0.50 | 0.15 | 0.23 | 20 |
| | Lgbtphobia | 0.78 | 0.75 | 0.76 | 171 |
| | Political | 0.61 | 0.53 | 0.57 | 220 |
| | Racism | 0.39 | 0.42 | 0.40 | 62 |
| | Religious intolerance | 0.27 | 0.16 | 0.20 | 19 |
| | Misogyny | 0.67 | 0.63 | 0.65 | 324 |
| | Xenophobia | 0.39 | 0.22 | 0.28 | 78 |
| | Other | 0.62 | 0.46 | 0.53 | 909 |
| | Not hate | 0.91 | 0.94 | 0.93 | 7177 |
| | **Micro avg** | **0.87** | **0.85** | **0.86** | **9065** |
| | **Macro avg** | **0.59** | **0.44** | **0.48** | |
| | **Weighted avg** | **0.85** | **0.85** | **0.85** | |
| | **Samples avg** | **0.87** | **0.86** | **0.86** | |

BERTimbau Base encounters difficulties in accurately identifying ageism, aporophobia, capacitism, and religious intolerance, resulting in lower precision, recall, and F1 scores. In contrast, the model excels in categories such as body shame, political, LGBTphobia, misogyny, xenophobia, and other, maintaining a delicate balance between precision and recall. Notably, racism poses a significant challenge for the base model, displaying subpar performance with a low support value of 62.

BERTimbau Large demonstrates improvements over the base model, particularly in categories with elevated precision, recall, and F1 scores. However, it grapples with challenges in accurately classifying instances of ageism, aporophobia, capacitism, and religious intolerance, as reflected in varying support values. While the larger model shows a modest enhancement in detecting racism, it still encounters hurdles in this domain, as indicated by relatively lower precision, recall, and F1 scores, alongside the low support.

Both models consistently identify instances that do not constitute hate speech, as evidenced by high precision, recall, and F1 scores for the "not hate" category, supported by a substantial overall support value.

However, persistent challenges arise in treating categories with limited training examples, as indicated by the variation in support values across different categories. In summary, while hate speech detection models, especially BERTimbau Large, demonstrate remarkable overall performance, there remains an opportunity for improvement, specifically in addressing the challenges associated with specific categories of hate speech and mitigating the categories in which annotated data is scarce. Improvement efforts and a nuanced approach to addressing imbalances in training data could significantly contribute to further advances in hate speech detection capabilities.

## 6. Conclusions

This investigation was driven by the goal of building the most significant public annotated dataset for hate detection in Brazilian Portuguese, and the resulting TuPy-E dataset has proven to be an accurate tool. Its substantial volume of meticulously curated instances guided our analyses and offered crucial insights into the intricate profile of hate speech in Brazil.

Throughout the examination, there was a notable focus on insults related to gender and political ideologies and enlightening intersections between the categories of misogyny, body shame, and LGBTphobia.

The N-gram analysis highlighted the fundamental role of context in deciphering the meanings of tweets. This analysis revealed a dual nature in using specific phrases, even in non-hateful contexts, which suggests a trivialization of specific hostile terms in neutral contexts.

The integration of graphs, a visual representation informed by the dataset, enriched our understanding, accentuating the prevalence of gender-based insults and political biases in the Brazilian hate speech scenario.

The graphical representation of hate speech categories provided a comprehensive view of how different types of hate co-occur, emphasizing the need for more instances, especially in specific categories, to improve the models' performance. This visualization highlighted intricate relationships between categories, offering a nuanced understanding of the interconnected nature of various forms of hate speech.

Experiments with the BERTimbau model showed commendable performance, implying the effectiveness of the base model for practical applications in detecting hate speech.

Extending our analysis to the BERTimbau Large model, we observed notable improvements, particularly in categories with higher precision, recall, and F1 scores. Despite facing challenges in specific categories of hate speech, the larger model demonstrated improved capabilities, affirming the positive impact of model size in improving hate speech detection performance.

Furthermore, it is essential to note that all components of this study, including the TuPy-E dataset, models, graph, and N-gram generation, are open-source. This commitment to open-source practices encourages collaboration and transparency in advancing research into detecting hate speech. Collaborative efforts between dataset curation and model experimentation, combined with insightful graphical representation, serve as the foundation for continued advancements in mitigating the impact of hate speech in Brazilian Portuguese content.

## Data availability

Data will be made available on repository: https://huggingface.co/datasets/Silly-Machine/TuPyE-Dataset

## 8. Attachments

Table A.1 - Words and expressions used to acquire twitter data.

| Hate Speech Category | Words and expressions |
|---|---|
| ageism | “Velho burro” / “velho babão” / “velho nojento” / “velho safado” / “velho tarado” / “coroa folgosa” /“velho pra isso” / “não tenho idade” / “jovem burro” / “adolescente maconheiro” |
| aporophobia | “Bolsa esmola” / “pobraiada” / “bandido favelado” / “favelado” / “filho de papai” / “pobre favelado” / “coisa de pobre” / “parece favelado” / “mendigo fedido” / “trabalhar vagabundo” |
| body shame | “Narigud”/ “gordo” / "gorda" / "baleia" / "doente" / "obeso" / "obesa" / "feia / "feio" / “gordo fazendo gordice” / “cabelo ruim”/ “cabelo de bombril” / “gorda escrota”/ “gordo escroto” / “feia pra caralho”/ “anão de jardim”/ “cão chupando manga” |
| capacitism | “retardado mental” / “tem down” / “alejado” / “demente” / “leproso” / “aidético” / “coisa de retardado” / “deficiente mental” / “é autista” / “parece cego”/ “boca aberta”/ “imbecil” /“esquizofrênic”/ “psicopata” |
| LGBTphobia | “boiola” / “baitola” / “cara de traveco” / “voz de traveco” / “queima rosca” / “meio afeminado” / “coisa de boiola” / “parece uma bixa” / “jeitinho de gay “/ “jeito de gay” |
| political | "esquerdista" / "petralha" / “petralha safada” / “coxinha burro” / “comunista safado” / “coxinha fascista” / “comunista”/ “político ladrão” / “compra votos” / “petista vagabundo” / “elite golpista” / “esquerda caviar” / "fascista" / "bolsonarista"/ "bolsominion" / "minion" / "nazista" / "lula" / "dilma" / "gulag" / "micheque" / "carluxo" / "milico" / "moro" / "ditador" / "patriota" / "liberal" |
| racism | “Cabelo ruim”/ “cabelo de bombril” / “não sou tuas nega” / “tinha que ser preto” / “da cor do pecado” / “preto é foda” / “nego é foda” / “cara de macaco”/ “preto safado”/ “negro fedido” / "macaco" / "mulata" / "moreno" / "morena" / "criolo" / "neguinho" / "neguinha" |
| religious intolerance | “crente do rabo quente” / “crente do cu quente” / “odeio crente” / “sem Deus no coração” / “bando de crente” / “tudo terrorista” / “padre pedofilo” / crente safado / crentona / “chuta que é macumba” |
| misogyny | “vadia”/ “safada” / “mal comida” / “coisa de mulherzinha“ / “falta de rola” / “falta de pica”/ “cara de puta” / “odeio mulher” / “feminazi” / “tinha que ser mulher” / "gorda" / "feia" / "vagabunda" / "piranha" / "escrota" / "maluca" / "feminista" / "puta" / "vaca" / "mal amada" |
| xenophobia | "povo burro” / “nordestino vagabundo” / “muçulmano bomba” / “baianice” / “baianada” / “tudo terrorista” / “volta pra sua terra” / "caiçara folgado" / "caipira burro" / "povo da roça" / "gaucho" |

Table A.2 - Count of the most frequent words of each hate category of TuPy-E dataset.

| Category | Word | Frequency |
|---|---|---|
| Misogyny | Mulher | 338 |
| | Feia | 169 |
| | Gorda | 167 |
| | Burra | 163 |
| | Sapatão | 118 |
| Political | Brasil | 101 |
| | Lula | 99 |
| | PT | 78 |
| | Povo | 64 |
| | L | 47 |
| LGBTphobia | Sapatão | 132 |
| | Viado | 83 |
| | Fufas | 81 |
| | Gay | 61 |
| | Mulher | 44 |
| Xenophobia | Refugiados | 62 |
| | Brasil | 24 |
| | Islão | 14 |
| | Povo | 11 |
| | Porra | 11 |
| Racism | Racismo | 24 |
| | Cara | 12 |
| | Cu | 12 |
| | Porra | 11 |
| | Vão | 10 |
| Body shame | Gorda | 156 |
| | Feia | 155 |
| | Cara | 27 |
| | Mulher | 20 |
| | Tão | 15 |
| Religious intolerance | Deus | 13 |
| | Islão | 13 |
| | Cristãos | 8 |
| | Diz | 7 |
| | Cristão | 6 |
| Capacitism | Cabeça | 8 |
| | Igual | 6 |
| | Cara | 6 |
| | Dedos | 4 |
| | Lula | 4 |
| Aporophobia | Pobre | 18 |
| | Fudido | 13 |
| | Brasil | 5 |
| | Lula | 5 |
| | Escola | 4 |
| Ageism | Velho | 11 |
| | Velha | 7 |
| | Veia | 6 |
| | Ódio | 5 |
| | Parece | 4 |
| Other | Cu | 228 |
| | Caralho | 167 |
| | Cara | 152 |
| | Porra | 135 |
| | Lixo | 102 |